\documentclass[letterpaper, 10 pt, conference]{ieeeconf}  % Comment this line out if you need a4paperRyan Razani

\IEEEoverridecommandlockouts                              % This command is only needed if 
\usepackage{graphics} % for pdf, bitmapped graphics files
\usepackage{epsfig} % for postscript graphics files
\usepackage{times} % assumes new font selection scheme installed
\usepackage{todonotes}
\usepackage{amsmath, amssymb, amsfonts}
\usepackage{subcaption}
\usepackage{float}
\usepackage{multirow}
\usepackage{wrapfig}
\usepackage{tabularx,multirow,booktabs,blindtext}
\usepackage{lmodern,babel,adjustbox,booktabs}
\usepackage{xcolor}
\usepackage{soul}
\title{\LARGE \bf
S3E-GNN: Sparse Spatial Scene Embedding with Graph Neural Networks for Camera Relocalization
}

\author{
Ran Cheng$^{1\ddagger*}$, Xinyu Jiang$^{1\ddagger}$, Yuan Chen$^{1}$, Lige Liu$^{1,2}$, Tao Sun$^{1, 2*}$
\thanks{$^1$ are with Lab2030, Midea Robozone Inc, Shanghai, 201704, China. ((e-mail: {chengran1, jiangxy77, chenyuan, liulg12, tsun}@midea.com).)}
\thanks{$^2$ are with School of Engineering, Massachusetts Institute of Technology, Cambridge, MA, 02139, USA. (e-mail: {xtllg, taosun}@mit.edu)}
\thanks{*Corresponding authors. $^{\ddagger}$Indicates equal contribution.}
}

\begin{document}

\maketitle
\thispagestyle{empty}
\pagestyle{empty}

%%%%%%%%%%%%%%%%%%%%%%%%%%%%%%%%%%%%%%%%%%%%%%%%%%%%%%%%%%%%%%%%%%%%%%%%%%%%%%%%
\begin{abstract}
Camera relocalization is the key component of simultaneous localization and mapping (SLAM) systems. This paper proposes a learning-based approach, named Sparse Spatial Scene Embedding with Graph Neural Networks (S3E-GNN), as an end-to-end framework for efficient and robust camera relocalization. S3E-GNN consists of two modules. In the encoding module, a trained S3E network encodes RGB images into embedding codes to implicitly represent spatial and semantic embedding code. With embedding codes and the associated poses obtained from a SLAM system, each image is represented as a graph node in a pose graph. In the GNN query module, the pose graph is transformed to form a embedding-aggregated reference graph for camera relocalization. We collect various scene datasets in the challenging environments to perform experiments. Our results demonstrate that S3E-GNN method outperforms the traditional Bag-of-words (BoW) for  camera relocalization due to learning-based embedding and GNN powered scene matching mechanism.

\end{abstract}

%%%%%%%%%%%%%%%%%%%%%%%%%%%%%%%%%%%%%%%%%%%%%%%%%%%%%%%%%%%%%%%%%%%%%%%%%%%%%%%%
\section{INTRODUCTION}
SLAM is the core technology for mobile robots to map their environments and localize themselves for navigation planning. It has been intensively studied in the robotics community in the past two decades \cite{174711, 4160954, 7219438, 7747236, Mur_Artal_2017, rosinol2020kimera}. Camera relocalization plays an important role in SLAM systems. It recovers the camera pose when the robot is in the state of tracking lost. It can also be applied in the loop closure detection to correct the accumulated drift in the trajectory to obtain a consistent map. Traditional camera relocalization methods rely on handcrafted features such as SIFT \cite{Lowe2004}, SURF \cite{bay2008speeded}, ORB \cite{ORB2011} and etc., and the bag-of-words (BoW) matching approaches \cite{angeli2008fast}, \cite{6094885}, \cite{galvez2012bags}. Due to the point-based characteristics, the handcrafted feature matching process may not be robust under challenging environments with conditions of low texture, repetitive patterns, illumination variation. In recent years, deep learning-based methods have become popular without the need of extracting and matching the handcrafted features. Kendall et al. \cite{kendall2015posenet} proposed PoseNet, which can regress 6-DOF camera pose quickly in an end-to-end fashion. However, there are issues of over-fitting and lack of generalization outside training datasets \cite{shavit2019introduction}. To overcome these limitations, researchers combined partial geometric constraints with deep learning-based methods, such as \cite{kendall2017geometric}, \cite{valada2018deep}, \cite{radwan2018vlocnet++}. These algorithms achieve state-of-the-art performance in existing indoor and outdoor datasets, but they only regress the initial camera pose estimation and rely on the results of sequence estimation. If the previous estimation is not accurate, errors will be accumulated, resulting in a system crash.

\begin{figure*}[!htb]
\vspace{5px}
    \centering
    \includegraphics[width=\textwidth]{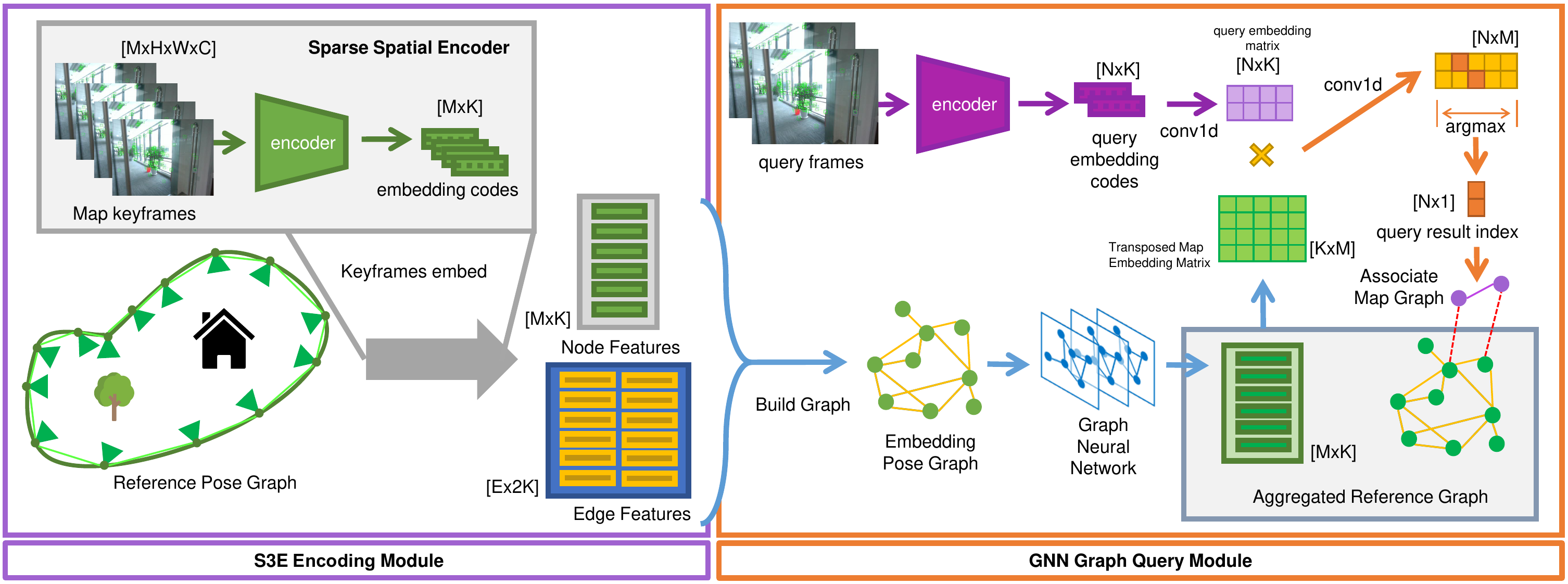}
    \caption{S3E-GNN pipeline for camera relocalization.}
    \label{fig:s3e_query_system_pipeline}
    \vspace{-10px}
\end{figure*}

To address the above issues, in this work, we propose an end-to-end learning framework named Sparse Spatial Scene Embedding with Graph Neural Networks (S3E-GNN) for efficient and robust camera relocalization, as shown in Fig. \ref{fig:s3e_query_system_pipeline}. The S3E-GNN pipeline has two modules. In the encoding module, the S3E network encodes images into embedding codes and generate a pose graph with the estimated poses. In the graph query module, a GNN transforms the pose graph to a reference graph where the features of the pose graph nodes are aggregated. To relocate the camera pose of a query image, we develop an efficient graph node querying algorithm based on calculating the inverse entropy multiplication between the embedding of the query image and the reference graph. To summarize, the main contributions of this paper are: 

\begin{itemize}
  \item A learning-based embedding network, S3E, encodes sensor measurements, as graph nodes.
  \item A GNN to transform a pose graph into a reference graph with aggregated features for each pose node. 
  \item A fast node query algorithm to locate the closest pose node in the reference graph to the query image.
\end{itemize}

To the best of our knowledge, this is the first work to combine a learning network for image embedding and a GNN for pose querying to perform camera relocalization. Our method does not need build a tree structure \cite{angeli2008fast} or use the nearest neighbors \cite{kendall2015posenet}, \cite{kang2021k} to search the BoW. In addition, we use a GNN query method rather than the CNN-based pose regression methods, \cite{kendall2015posenet}, \cite{sattler2019understanding}, \cite{xue2020learning} to improve the robustness of the SLAM system in challenging environments. 

\section{RELATED WORKS}
\label{sec:citations}

\subsection{Handcrafted feature-based methods} 

In 2D vision, an image is represented by the BoW method. The camera pose of the query image can be estimated by matching it with the most similar image in the image database. In 3D vision, a 3D scene map composed of point clouds can be generated from the Structure from Motion (SfM) \cite{sfm1999} with RGB-D images. Relocalizing 2D images in the 3D map can be achieved by finding the 2D-3D correspondences between 2D image pixels and 3D points. Shotton et al. \cite{shotton2013scene} developed scene coordinate regression forests to predict the correspondences between image pixels and 3D points in the 3D world frame. Recently, Nadeem et al. \cite{nadeem2019direct} trained a Descriptor-Matcher to directly find the correspondences between the 2D descriptors in the RGB query images and the 3D descriptors in the dense 3D point clouds to relocate the pose of the 6-DOF camera.

\subsection{Learning feature-based method}
Gao and Zhang \cite{gao2017unsupervised} first used the stacked denoising auto-encoder to automatically learn compressed representations from raw images in an unsupervised manner. Although no manual design of visual features is required, the network's effectiveness depends heavily on the tuning of hyperparameters. Kendall et al. \cite{kendall2015posenet} proposed PoseNet to regress 6-DOF camera pose. Although the image embeddings from CNNs contain both high level information, the pose information, which is significant for data association, is not stored. GNNs benefit from their graph structure, which can pass information between nodes through edges, making them more suitable for handling relational data. Recently, the use of GNNs in close-loop detection \cite{cascianelli2017robust, yue2019robust}, camera relocalization \cite{xue2020learning, elmoogy2021pose}, etc. has been developing rapidly. For camera relocalization, Xue et al. \cite{xue2020learning}  extracted image features using CNN, and used the extracted feature map as graph nodes to build the graph by joint GNN and CNN iterations. Elmoogy et al. \cite{elmoogy2021pose} used ResNet50 for feature extraction of the input image and used the vectors flattened from the extracted feature maps as graph then obtained the camera pose by GNN.

%========================================================================

\section{METHODS}
\label{sec:methods}

Our S3E-GNN framework is composed of two modules: the S3E encoding module and the GNN query module.

\subsection{S3E Encoding Module}

\begin{figure}[htb]
    \vspace{5px}
    \centering
    \includegraphics[width=0.45\linewidth]{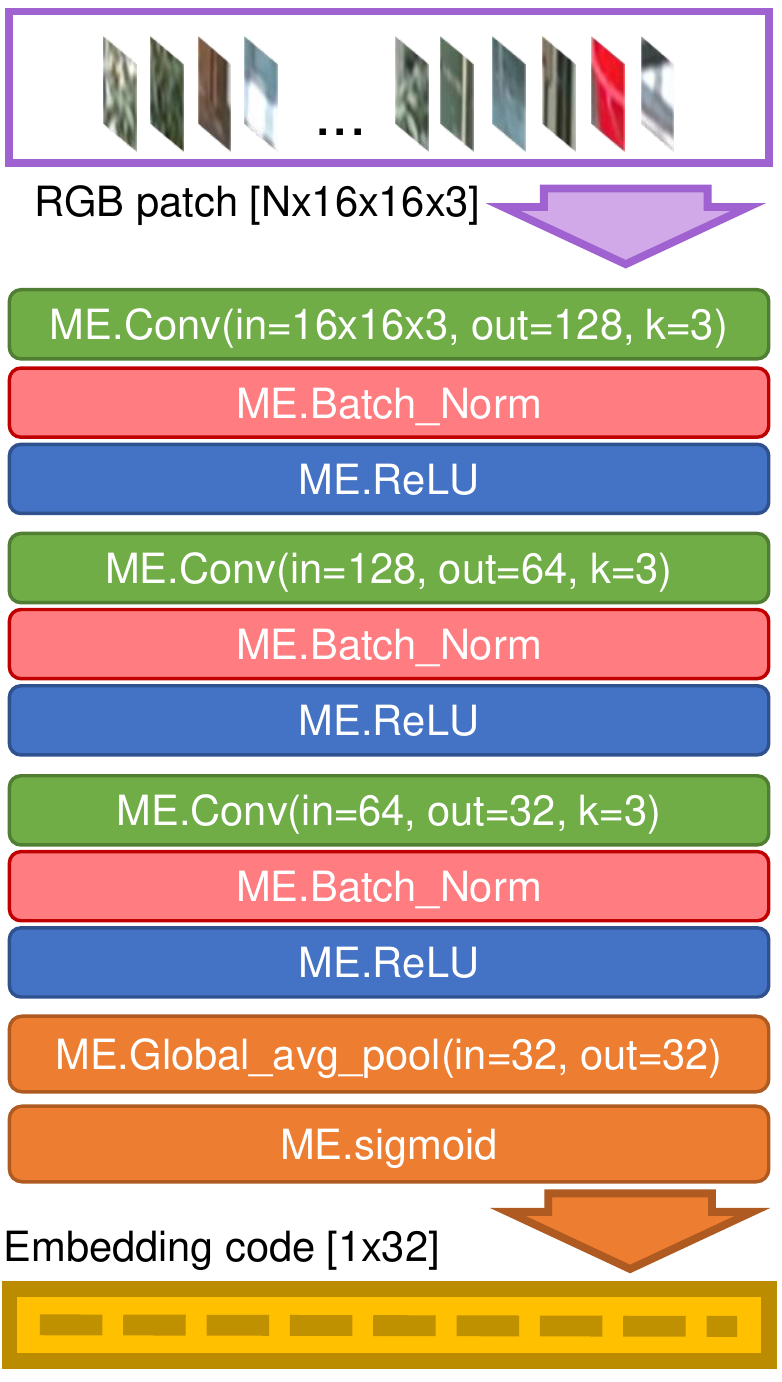}
    \caption{Encoder backbone network based on Sparse Convolution, ME means MinkowskiEngine \cite{choy20194d}.}
    \label{fig:encoder_arch}
\end{figure}

The S3E module encodes the an image into its embedding code. For encoding efficiency, we did not directly use the entire image as the input. Instead, we firstly use VINS-RGBD \cite{shan2019rgbd} SLAM system to randomly select 128 ORB feature points in the image. Then, we sliced a 16x16 patch around each feature point. The total 128 pieces of patches together with their pixel coordinates, which contain spatial information, as the representation of the original image are fed into the encoder. We develop two types of S3E encoder backbones. One used Visual Transformer (ViT) \cite{dosovitskiy2020image} and the other, as shown in Fig. \ref{fig:encoder_arch}, is based on sparse convolution (SparseConv) \cite{choy20194d}. SparseConv encoder can take arbitrary data size, while ViT encoder requires a fixed number of patches. As shown in Fig. \ref{fig:encoder_arch}, the RGB patches taken from the image are stacked and fed into SparseConv encoder built from MinkowskiEngine \cite{choy20194d}. The input RGB patches and the corresponding 2D pixel coordinates are converted into a sparse tensor. Note that the feature points on the image is too sparse so that the sparse convolution kernel is hard to capture neighborhood, therefore we down-sample the tracked feature points' pixel coordinates by patch size. The encoder consists of three sparse convolutions (with batch norm and ReLU) followed by a global average pooling layer to aggregate all the patch information into a single embedding code to represent the image.

Next, we build the embedding dataset to train the encoder. The ground truth data for training is generated by following the below steps, shown in Fig. \ref{fig:embedding_dataset_pipeline}:

\begin{figure}[htbp]
    \centering
    \includegraphics[width=\linewidth]{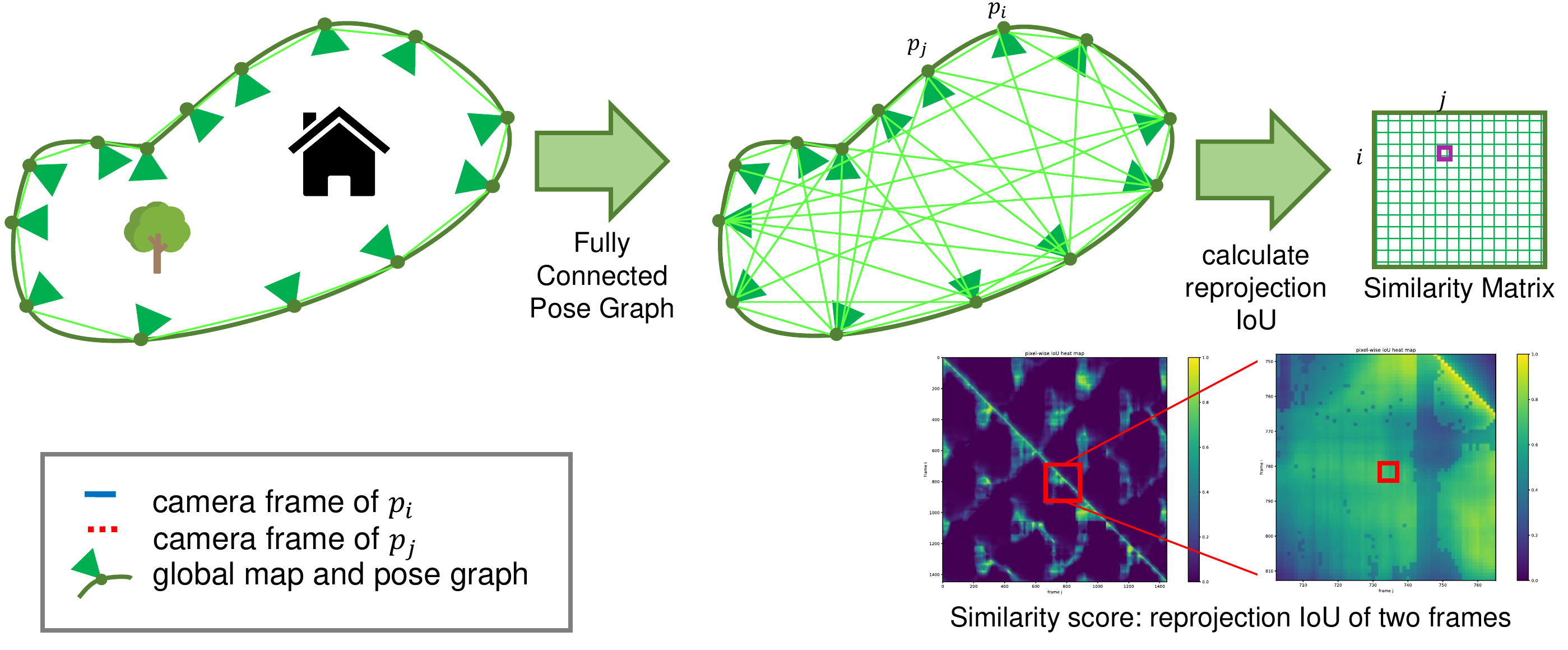}
    \caption{Embedding dataset build pipeline.}
    \label{fig:embedding_dataset_pipeline}
    \vspace{-10px}
\end{figure}

\begin{enumerate}
    \item Estimate camera keyframe pose graph using VINS-RGBD \cite{shan2019rgbd}. The estimated camera poses form a pose graph, in which each node is a keyframe. 
    \item Build the fully connected graph from the keyframe nodes and calculate the reprojection Intersection over Union (IoU) for every pair of the keyframes.
    \item Establish the similarity matrix as the ground truth data for training the encoder.
\end{enumerate}

\begin{figure}[htbp]
    \centering
    \includegraphics[width=\linewidth]{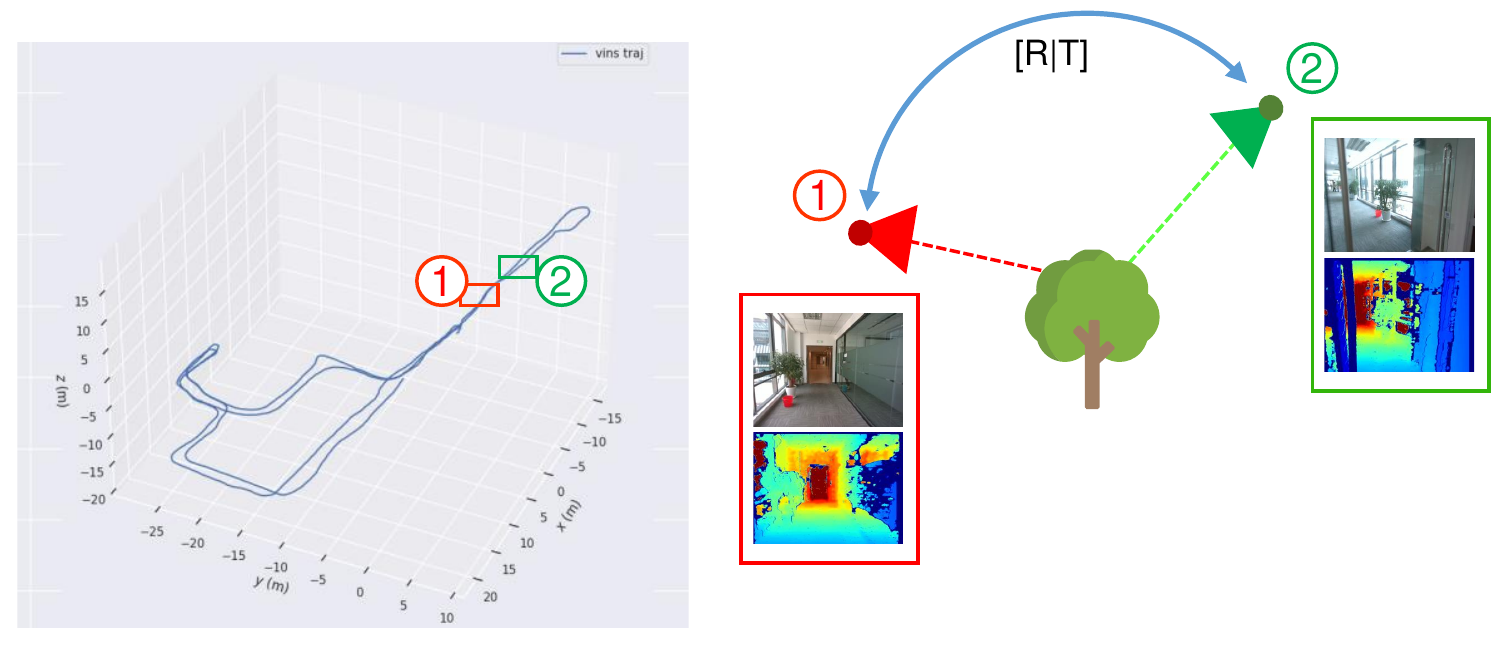}
    \caption{Reprojection IoU calculation illustration.}
    \label{fig:reproj_iou}
    \vspace{-10px}
\end{figure}

\begin{equation}
    s(p_i, p_j) = \frac{\sum_{n \in N}{\pi(K\xi_j\xi_i^{-1}(K^{-1}(x_n, z_n)))}}{N}
    \label{eq:reproj_iou}
\end{equation}

To calculate the IoU between two keyframes: the source and the target, we count the number of projected pixels that are from the source keyframe inside the target keyframe and divide it by the keyframe resolution to obtain the normalized similarity score. Fig. \ref{fig:reproj_iou} and Equation. \ref{eq:reproj_iou} described the reprojection IoU calculation process. Function $\pi$ is the validation function that detects if the input coordinate is inside the target keyframe. $\xi$ is the camera pose, $N$ is image resolution, $K$ is the camera intrinsics. $x_n$ is $n$th pixel coordinate. $z_n$ is the depth taken from depth map .  Since the reprojection IoU implicitly incorporates the camera intrinsics and depth information, the encoder can learn geometric information. This property advantages the robustness of the GNN query in the next section. The similarity scores of every pair of keyframes form the similarity matrix as shown in Fig. \ref{fig:embedding_dataset_pipeline}, which quantitatively evaluates the similarity across keyframes.

\begin{figure}[htb]
    \centering
    \includegraphics[width=\linewidth]{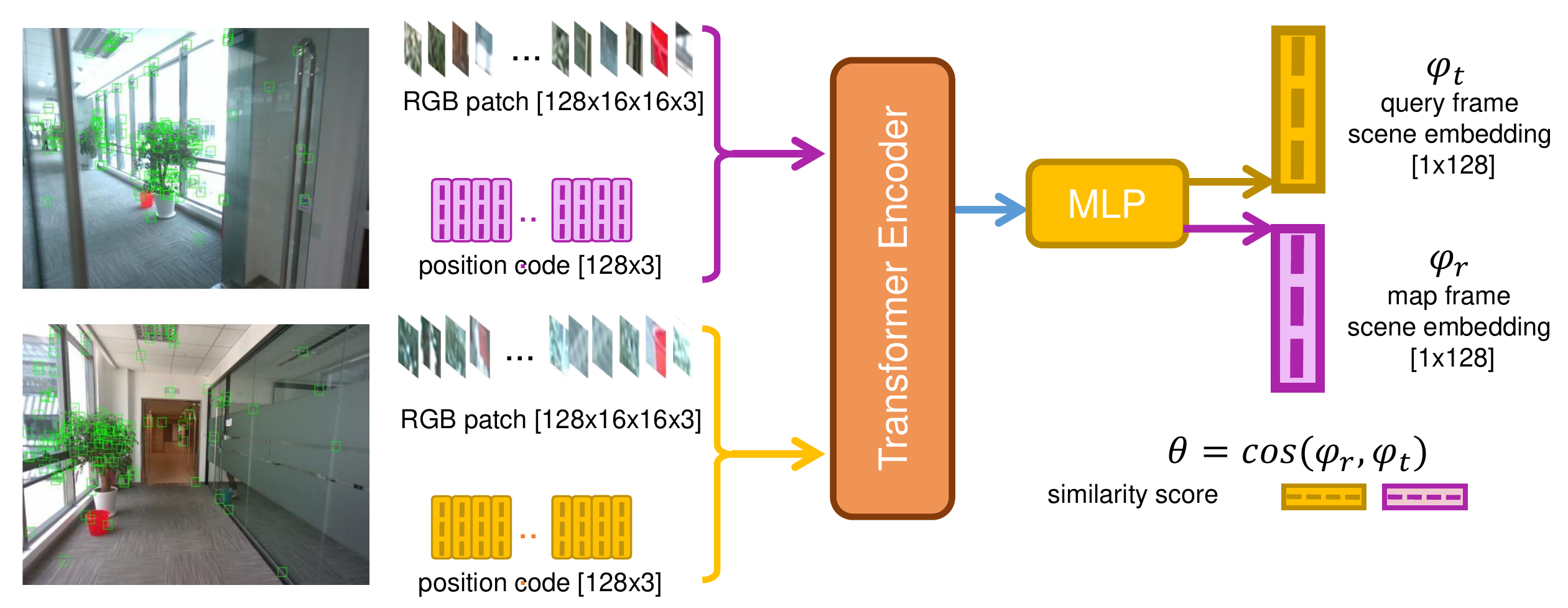}
    \caption{Embedding network training pipeline.}
    \label{fig:embedding_net_arch}
\end{figure}

With the similarity matrix from the IoU calculation as the ground truth data, we can train the encoder. Fig. \ref{fig:embedding_net_arch} shows that the encoder converts the input RGB patches of two keyframes into their latent embeddings. The cosine similarity is applied to the two latent embeddings to output the similarity score from the encoder. The cost function for backpropagation is defined as the shrinkage loss \cite{lu2018deep} between the similarity scores from the encoder output and the ground truth data. After the loss converges, the trained encoder can be applied to a new query keyframe to generate its embedding code.

\subsection{GNN Query Module}

The GNN query module transforms the embedding pose graph into a feature aggregated reference graph, which is used for querying a sub-graph composed of successive keyframes to find the most matching nodes for camera relocalization.

\begin{figure}[htb]
    \centering % <-- added
\begin{subfigure}{0.23\textwidth}
  \includegraphics[width=\linewidth]{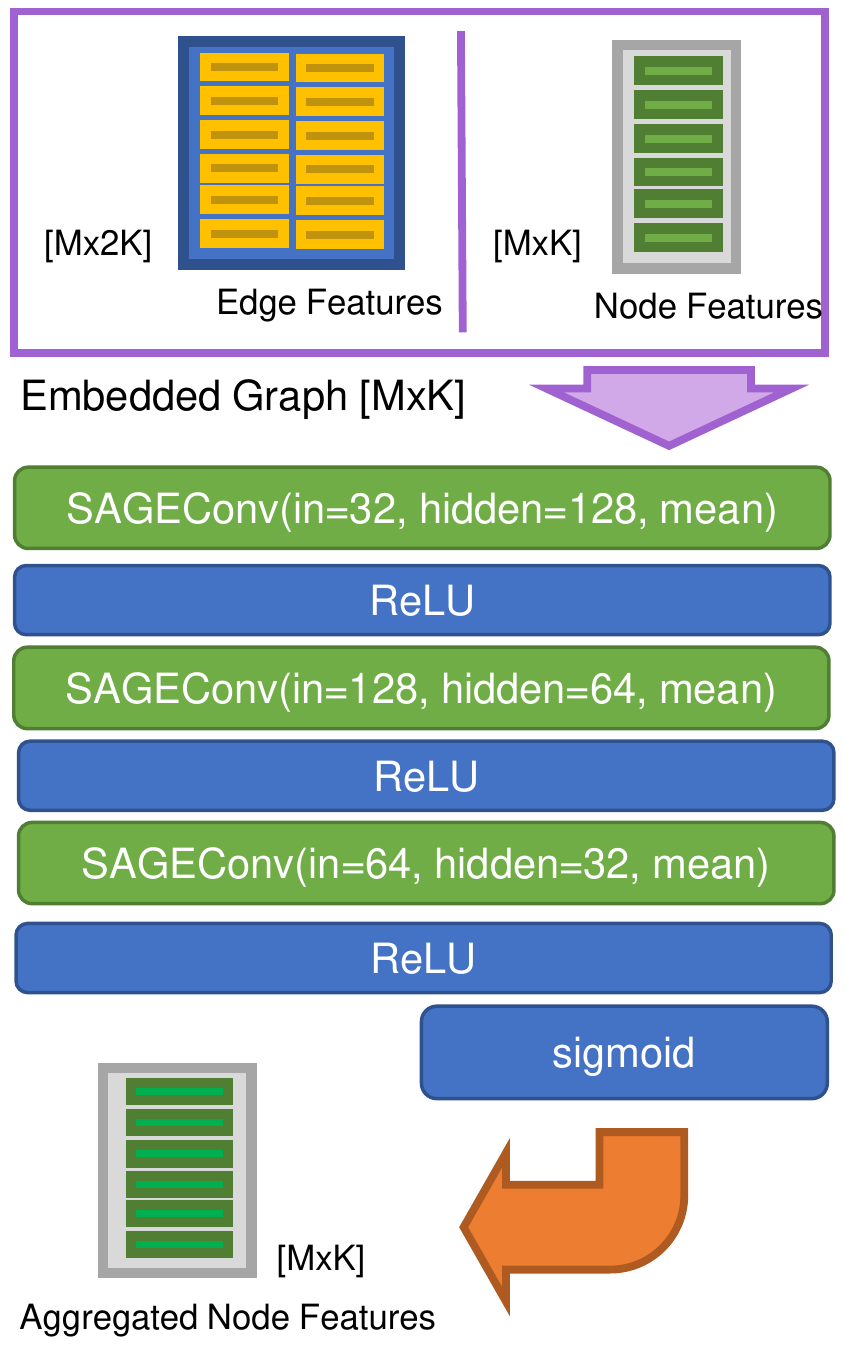}
  \caption{GNN for map graph and query sub-graph.}
  \label{fig:gcn_arch}
\end{subfigure}
\begin{subfigure}{0.18\textwidth}
  \includegraphics[width=\linewidth]{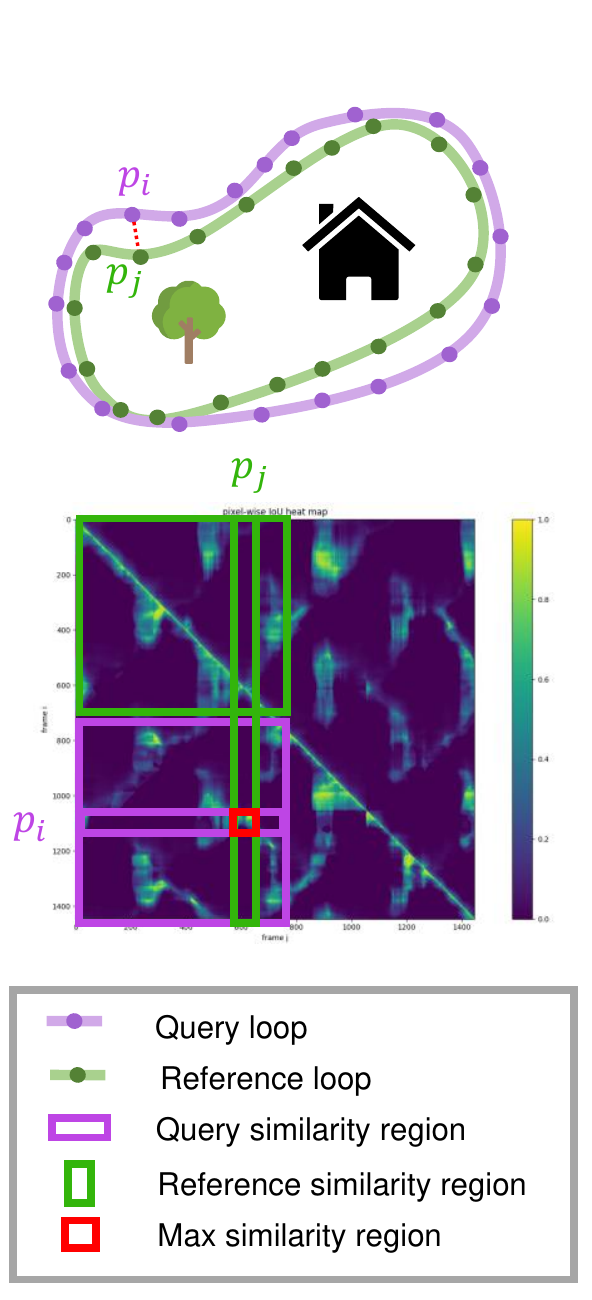}
  \caption{Ground truth label preparation pipeline for GNN.}
  \label{fig:train_gcn}
\end{subfigure}\hfil % <-- added
\caption{}
\label{fig:evaluation_scene}
\vspace{-10px}
\end{figure}

% \begin{figure}[htb]
%     \centering
%     \includegraphics[width=0.5\linewidth]{figs/s3e_gcn_network_arch3.pdf}
%     \caption{GNN for map graph and query sub-graph.}
%     \label{fig:gcn_arch}
% \end{figure}

Once we have the embedding code of every keyframe, we concatenate the embeddings of two nodes connected by an edge as an edge embedding. The node and edge embeddings are used to build an embedding pose graph to feed into a GNN (as shown in Fig. \ref{fig:gcn_arch}), which is comprised of three layers of SAGEConv layers \cite{hamilton2017inductive} followed by ReLU and finally is clipped with a Sigmoid function. To train the GNN, we use the similarity matrix from the reprojection IoU as the ground truth label for the GNN query results. First we take two runs of loop trajectories in the same scene. One loop is used as the reference loop. The other is used as as the query loop. As shown in Fig. \ref{fig:train_gcn} The ground truth match logits for frame $i$ in the query sub-graph and the reference map is determined by selecting the corresponding columns (reference map loop region) in $i$th row in the reprojection IoU matrix. The cost function is defined as the cross entropy of predicted probability of graph query network and the corresponding row in reprojection IoU map.

With a trained GNN, a feature aggregated reference graph is obtained from the embedding pose graph. This reference graph replaces the BoW scheme for camera relocalization. To improve the precision and robustness of relocalization, we use a group of successive keyframes rather than a single keyframe to form a query sub-graph. Since the query sub-graph and the reference graph have the same embedding dimension, we can apply an inverse cross-entropy matrix multiplication between the embedding matrices of the query sub-graph and the reference graph to produce a similarity matrix and detect a loop closure by finding the maximum value in the similarity matrix. The inverse cross-entropy multiplication is defined as below:

\begin{equation}
    U_{ij} = \frac{1}{\eta -\sum_{k \in K}{q_{ik}log(m_{kj})}}
\end{equation}

which applies the row-column wise inverse cross-entropy instead of aggregating the row-column multiplication result directly, where $U$ is the output of the inverse cross-entropy multiplication, $K$ is the embedding dimension, $q$ and $m$ are query sub-graph embedding matrix and reference graph embedding matrix. For each element of $U$, $U_{ij}$ is calculated by the inverse cross-entropy of $i$th row of query embedding matrix and $j$th column of reference graph embedding matrix.
 
With the multiplication results, we use graph optimization algorithms \cite{grisetti2011g2o} to optimize the global poses. For a loop closure detection pair: query frame $p_i$ and reference frame $p_j$, the graph optimization process is as below:

\begin{equation}
    T_w^i = \underset{T_w^*}{\mathrm{argmin}}\, r_{i, j}^T\Omega_{i, j} r_{i, j}
\end{equation}

\begin{equation}
    r_{i, j} = log(T^{j}_{i}T^{w}_{j}T^{i}_{w})
\end{equation}

Where $T_w^i$ is the camera pose of frame $p_i$. We initialize it with its previous tracked pose $T_w^{i-1}$, $\Omega = \hat{\theta} \times I$ is the information matrix prior, which is scaled by the embedding similarity vector $\hat{\theta}$. All the poses in the graph are adjusted to minimizing the residual between reference frame and query frame: $r_{i, j}$.

Since we sparsify the keyframes by cropping tracked feature points neighborhood, and use inverse cross-entropy matrix multiplication to replace the BoW database and matching procedure, the overall performance impact of this method on a SLAM system is marginal.

%===============================================================================

\section{EXPERIMENTS}

\subsection{Dataset Collection and Processing}

\textbf{Scene data collection:} We record multiple scenes in our office with several sequences to ensure at least two complete cycles using Intel Realsense D455i. Note that we do not use open-source datasets since we find that none of them have multiple loops and ground-truth trajectory to compute the IOU and high-precision depth. Our dataset provides RGB images, depth images, and IMU data. 

\textbf{Localization data generation:} We first use the VINS-RGBD system \cite{shan2019rgbd} to reconstruct the scene using the recorded video data. After generate the scene map, we save the camera pose trajectory, keyframes and extracted patches around feature points. To increase the scalability, we extract patches at multiple scales: $16 \times 16$, $32 \times 32$, $64 \times 64$. 

\begin{figure}[htb]
  \centering
  \captionsetup{font=small,labelfont=small}
 {\includegraphics[trim=0pt 0pt 0pt 0pt,width=0.43\linewidth]{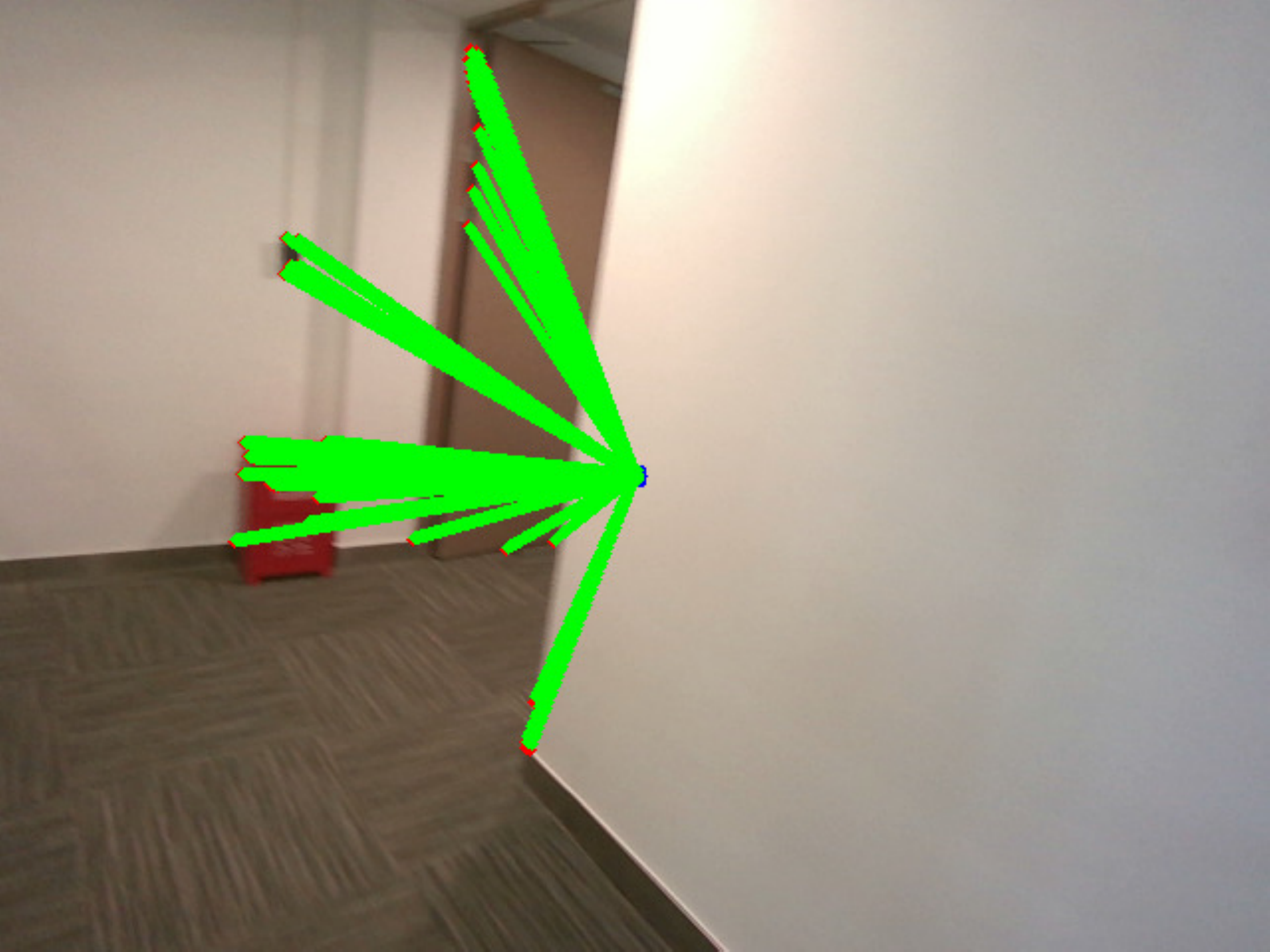}\label{fig:map_neighbor}} \hspace{2px}
  {\includegraphics[trim=0pt 0pt 0pt 0pt,width=0.43\linewidth]{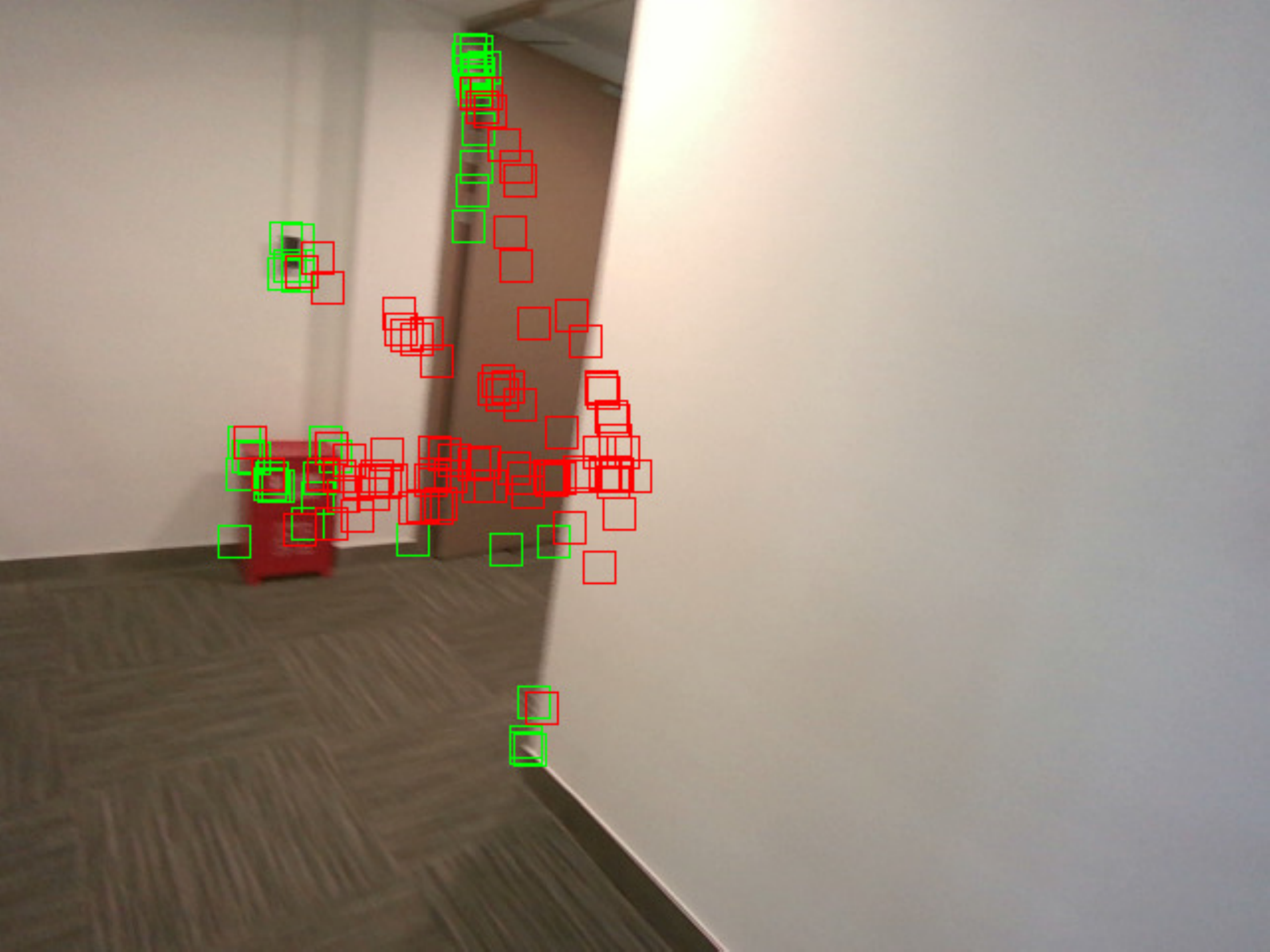}\label{fig:crash_map}}
  \caption{Line extraction and random patch selection overview}
   \label{fig:patch_selection}

\end{figure}

\textbf{Low texture data augmentation:} During the scene data collection, we may encounter some low texture environments. We propose a feature completion strategy to supplement the features in the low texture environments. If the number of features is below 128, we find all the lines from each key point to the image center, and then randomly select the points on the lines and extract the patches. Fig. \ref{fig:patch_selection} shows the process of constructing lines using existed feature points and random patch selection.

\setlength{\columnsep}{4pt}%
\vspace{10px}

\textbf{Hidden surface removal:} When we calculate the IoU by projecting keyframe A to keyframe B. If the projected points are behind the scene in keyframe B, they should not be visible. However, without incorporating the depth information, these points can be mistakenly projected onto keyframe B. To correct this hidden surface issue, we apply the z-Buffer algorithm \cite{10.1145/166117.166147} to handle the depth relation and remove the hidden surfaces from the projected image.

\subsection{Quantitative Evaluation:} 

We perform our experiments on several challenging environments, including low texture, dynamic objects, and strong illumination variations. The total dataset contains more than 164,380K keyframes and 3,230K loop frames. We apply SMOTE up-sampling \cite{Chawla2002SMOTESM} to balance the loop and non-loop frames. We also incorporate random flips on the patch images. We collect 5 different scenes and use 2 for the training sequences.2 for the test sequences and 1 for the validation sequence. Due to the hardware limitation of the camera, we only perform our experiments in an indoor environment. We design the trajectory path by collecting the same scene with two reversed loops so that our S3E-GNN method can generate extra edges as constraints compared with the traditional BoW methods.

We train our two S3E backbones: ViT and SparseConv with the same hyper-parameters: $0.001$ learning rate, $10^{-5}$ weight decay rate, 0.95 momentum, we adopt cyclic scheduler with max learning rate as $0.1$, and SGD optimizer for 140 epochs. We compare the run-time speed and performance of these two backbones and the effect of GNN in loop closure detection using VINS-Mono SLAM \cite{qin2018vins}. Table. \ref{table:ablation_table} demonstrates the comparison of the performance boost after incorporating the GNN query module. For the GNN, since the network prone to diverge, we propose a lower learning rate: $10^{-5}$, the rest of the hyper-parameters are the same as S3E backbones. LC is the loop closure graph optimization backend. The pose estimation error drops drastically with ViT powered S3E-GNN while the run-time speed drops as well. The SparseConv powered S3E-GNN achieves less improvement yet maintains higher run-time speed. This is because the SparseConv network only convolves the immediate neighborhood patches, resulting in lacking global context, compared to ViT.

\begin{table}[H]

\captionsetup{font=scriptsize,labelfont=scriptsize}
\caption{Ablation study of proposed method vs baseline (VINS-Mono), LC means loop closure.}
\label{table:ablation_table}
\adjustbox{max width=\columnwidth}{
\begin{tabular}{lllllll}
\hline
\hline
Architecture & S3E (ViT) & S3E (SparseConv) & LC  & GNN+LC & RMSE (RMSE)   & FPS  \\ \hline
Baseline     &           &                  &     &        & 3.73 (0.45\%) & 15.3 \\ \hline
Proposed     & 16x16     &                  & $\checkmark$ &        & 3.77 (0.38\%) & 13.9 \\
             & 32x32     &                  & $\checkmark$ &        & 3.73 (0.53\%) & 9.1  \\
             & 64x64     &                  & $\checkmark$ &        & 3.37 (0.54\%) & 5.7  \\
             &           & $\checkmark$              & $\checkmark$ &        & 3.71 (0.61\%) & 14.3 \\
             & 16x16     &                  &     & $\checkmark$    & 3.89 (0.38\%) & 12.1 \\
             & 32x32     &                  &     & $\checkmark$    & 3.15 (0.41\%) & 8.3  \\
             & 64x64     &                  &     & $\checkmark$    & 2.74 (0.56\%) & 3.8  \\
             &           & $\checkmark$              &     & $\checkmark$    & 3.13 (0.31\%) & 13.7 \\ \hline \hline
\end{tabular}
}
\vspace{-10px}
\end{table}

\begin{table}[htbp]
\captionsetup{font=scriptsize,labelfont=scriptsize}
\caption{Pose error compared with ground truth trajectory. RMSE of ATE (Absolute Translational Error) is in cm, and is averaged over all test datasets (Office02, Office03, Home03)}
\label{table:evaluation_table}
\adjustbox{max width=\columnwidth}{
\begin{tabular}{|lllllll|}
\hline
Methods                    & IMU                  & RGBD                 & S3E/BoW & RMSE & $\sigma$RMSE & RMSE Max \\ \hline
\multirow{2}{*}{ORB-SLAM3 \cite{campos2021orb}} & \multirow{2}{*}{$\checkmark$} & \multirow{2}{*}{} & S3E-GNN & \textbf{1.98} & \textbf{0.82}\% & \textbf{12.33} \\ \cline{4-7} 
                           &                      &                      & BoW     & 3.15  & 0.8\%  & 20.32    \\ \hline
\multirow{2}{*}{VINS-Mono\cite{qin2018vins}} & \multirow{2}{*}{$\checkmark$} & \multirow{2}{*}{}    & S3E-GNN     & \textbf{1.33}  & \textbf{0.31}\% & \textbf{11.71}    \\ \cline{4-7} 
                           &                      &                      & BoW     & 3.19  & 1.0\%  & 17.80    \\ \hline
\multirow{2}{*}{VINS-RGBD\cite{shan2019rgbd}} & \multirow{2}{*}{$\checkmark$}    & \multirow{2}{*}{$\checkmark$} & S3E-GNN     & \textbf{1.19 } & \textbf{0.4}\%  & \textbf{11.13}    \\ \cline{4-7} 
                           &                      &                      & BoW     & 3.73  & 0.4\%  & 25.43    \\ \hline
\multirow{2}{*}{S-MSCKF \cite{geneva2019linear}}   & \multirow{2}{*}{$\checkmark$} & \multirow{2}{*}{}    & S3E-GNN     & \textbf{2.82}  & \textbf{8.8}\%  & \textbf{13.21}    \\ \cline{4-7} 
                           &                      &                      & BoW     & 6.32  & 9.2\%  & 32.05    \\ \hline
\multirow{2}{*}{RGBDTAM \cite{concha2017rgbdtam}}   & \multirow{2}{*}{}    & \multirow{2}{*}{$\checkmark$} & S3E-GNN     & \textbf{3.35}  & \textbf{0.2}\%  & \textbf{9.78 }    \\ \cline{4-7} 
                           &                      &                      & BoW     & 9.98  & 0.6\%  & 11.21    \\ \hline
\end{tabular}
}
\end{table}

\begin{table*}[!htbp]
\vspace{5px}
\centering
\caption{Generalization experiment by only using inference module that training in ourselves office dataset: translation (m) and rotation ($^\circ$) error comparison in Microsoft 7-scenes dataset \cite{shotton2013scene}.}
\label{table:Microsoft7scene_table}
\begin{tabular}{ccccccccc}
\hline
\multirow{2}{*}{Method} & \multicolumn{8}{c}{Sequence}                                                                                     \\
                        & Chess       & Fire         & Heads        & Office      & Pumpkin     & Kitchen     & Stairs       & Avg         \\ \hline
PoseNet15\cite{kendall2015posenet}               & 0.32, 8.12 & 0.47, 14.4  & 0.29, 12.0  & 0.48, 7.68 & 0.47, 8.42 & 0.59, 8.64 & 0.47, 13.8  & 0.44, 10.4 \\
Hourglass\cite{melekhov2017image}               & 0.15, 6.17 & 0.27, 10.84 & 0.19, 11.63 & 0.21, 8.48 & 0.25, 7.01 & 0.27, 10.15 & 0.29, 12.46 & 0.23,9.53  \\
LSTM-Pose\cite{walch2017image}               & 0.24, 5.77 & 0.34, 11.9  & 0.21, 13.7  & 0.30, 8.08 & 0.33, 7.00 & 0.37, 8.83 & 0.40, 13.7  & 0.31, 9.85 \\
ANNet\cite{bui2019adversarial}                   & 0.12, 4.30 & 0.27, 11.60 & 0.16, 12.40 & 0.19, 6.80 & 0.21, 5.20 & 0.25, 6.00 & 0.28, 8.40  & 0.21, 7.90 \\
BranchNet\cite{wu2017delving}               & 0.18, 5.17 & 0.34, 8.99  & 0.20, 14.15 & 0.30, 7.05 & 0.27, 5.10 & 0.33, 7.40 & 0.38, 10.26 & 0.29, 8.30 \\
GPoseNet\cite{cai2019hybrid}                & 0.20, 7.11 & 0.38, 12.3  & 0.21, 13.8  & 0.28, 8.83  & 0.37, 6.94 & 0.35, 8.15 & 0.37, 12.5  & 0.31, 9.95  \\
MLFBPPose\cite{wang2019discriminative}               & 0.12, 5.82 & 0.26, 11.99 & 0.14, 13.54 & 0.18, 8.24 & 0.21,7.05  & 0.22, 8.14 & 0.38, 10.26 & 0.22, 9.29 \\
VidLoc\cite{clark2017vidloc}                  & 0.18, NA   & 0.26, NA    & 0.14, NA    & 0.26, NA   & 0.36, NA   & 0.31, NA   & 0.26, NA    & 0.25, NA   \\
MapNet\cite{gao2019graph}                  & 0.08, 3.25 & 0.27, 11.69 & 0.18, 13.25 & 0.17, 5.15 & 0.22, 4.02 & 0.23, 4.93 & 0.30, 12.08 & 0.21, 7.77 \\
LsG\cite{xue2019local}                     & 0.09, 3.28 & 0.26, 10.92 & 0.17, 12.70 & 0.18, 5.45 & 0.20, 3.69 & 0.23, 4.92 & 0.23, 11.3  & 0.19, 7.47 \\ \hline
S3E-GNN                    & 0.21, 6.12  & 0.45, 15.27 & 0.21, 11.7  & 0.19, 5.56 & 0.33, 5.12 & 0.28, 5.63 & 0.32, 13.1  & 0.28, 8.92 \\ \hline
\end{tabular}
\vspace{-10px}
\end{table*}

To benchmark with the traditional BoW loop closure detection,  we evaluate the five state-of-the-art SLAM methods with ViT powered S3E-GNN as shown in Table. \ref{table:evaluation_table}. All the methods with the S3E-GNN outperform the traditional BoW loop closure methods by a large margin. We evaluate the RMSE (Root Mean Square Error), $\sigma RMSE$ (standard deviation of RMSE) of the Absolute Translational Error compared with the ground truth trajectory averaged over 10 runs to ensure repeatability. To evaluate the generalization of our method, we test S3E-GNN that is trained on our office dataset on Microsoft 7-scenes dataset \cite{shotton2013scene} directly. As shown in Table. \ref{table:Microsoft7scene_table}, our work is on par with the state-of-art methods, which demonstrates reasonable out-of-dataset generalizability. Since S3E-GNN is trained on the our office dataset, we get the best result on the office scene of 7 scenes dataset.

\begin{figure}[htb]
    \centering
    \includegraphics[width=\linewidth]{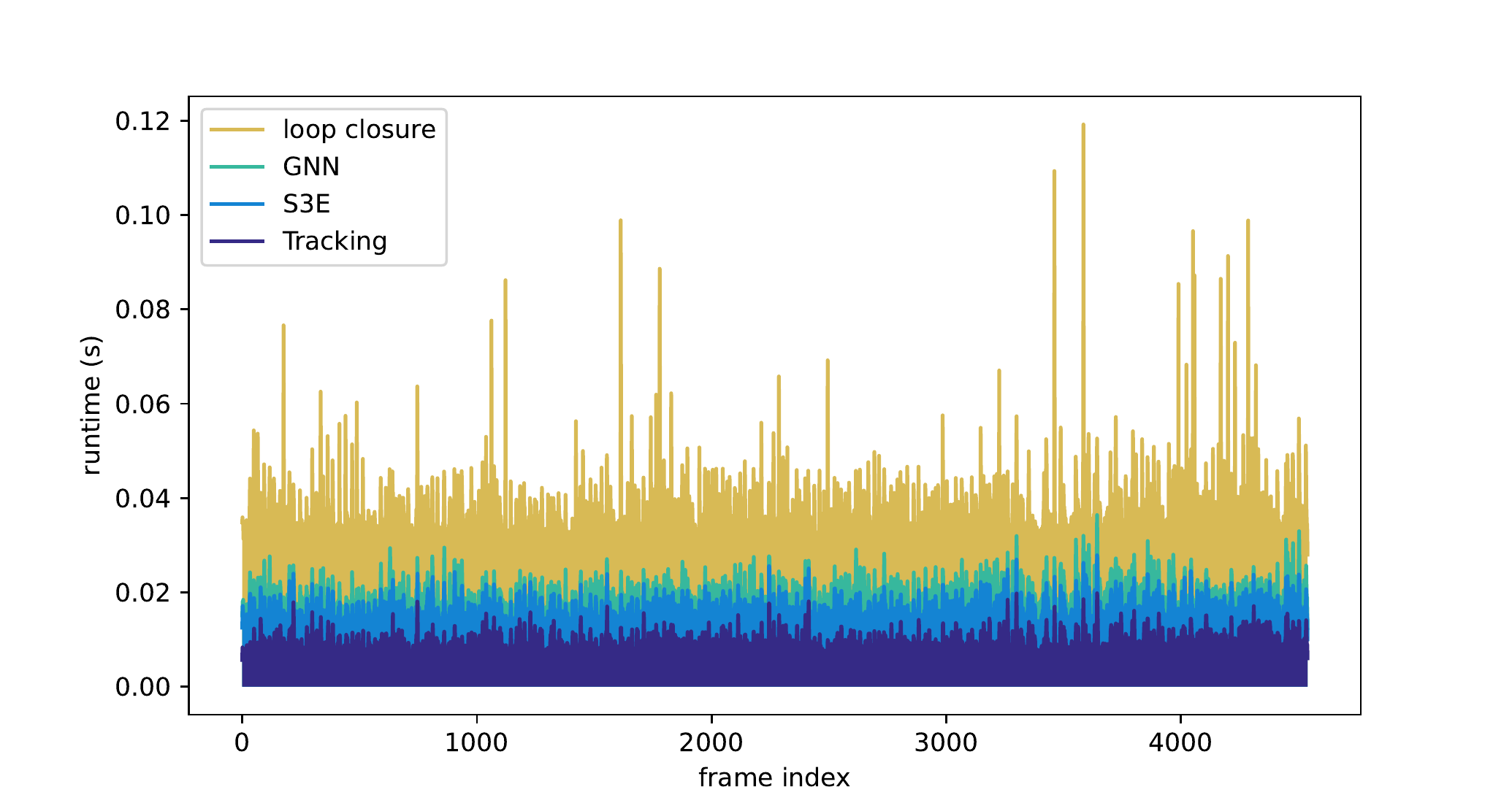}
    \caption{The CPU runtime of different components and the breakdown of the proposed S3E-GNN system.}
    \label{fig:runtime_eval}
\end{figure}

\textbf{Runtime evaluation:} We run the full system (VINS-Mono + S3E-GNN) on a i7 10700k CPU with RTX2080 GPU PC. Fig. \ref{fig:runtime_eval} shows the CPU single-round runtime of the full system running on a testing sequence. The time consumption of S3E and GNN component is marginal compared to the tracking and loop closure module. As additional edges were introduced from the GNN query results, graph optimization in the loop closure thread consumes the majority of the computation time.

\subsection{Qualitative Evaluation:}

\begin{figure}[htb]
    \centering % <-- added
\begin{subfigure}{0.15\textwidth}
  \includegraphics[width=\linewidth]{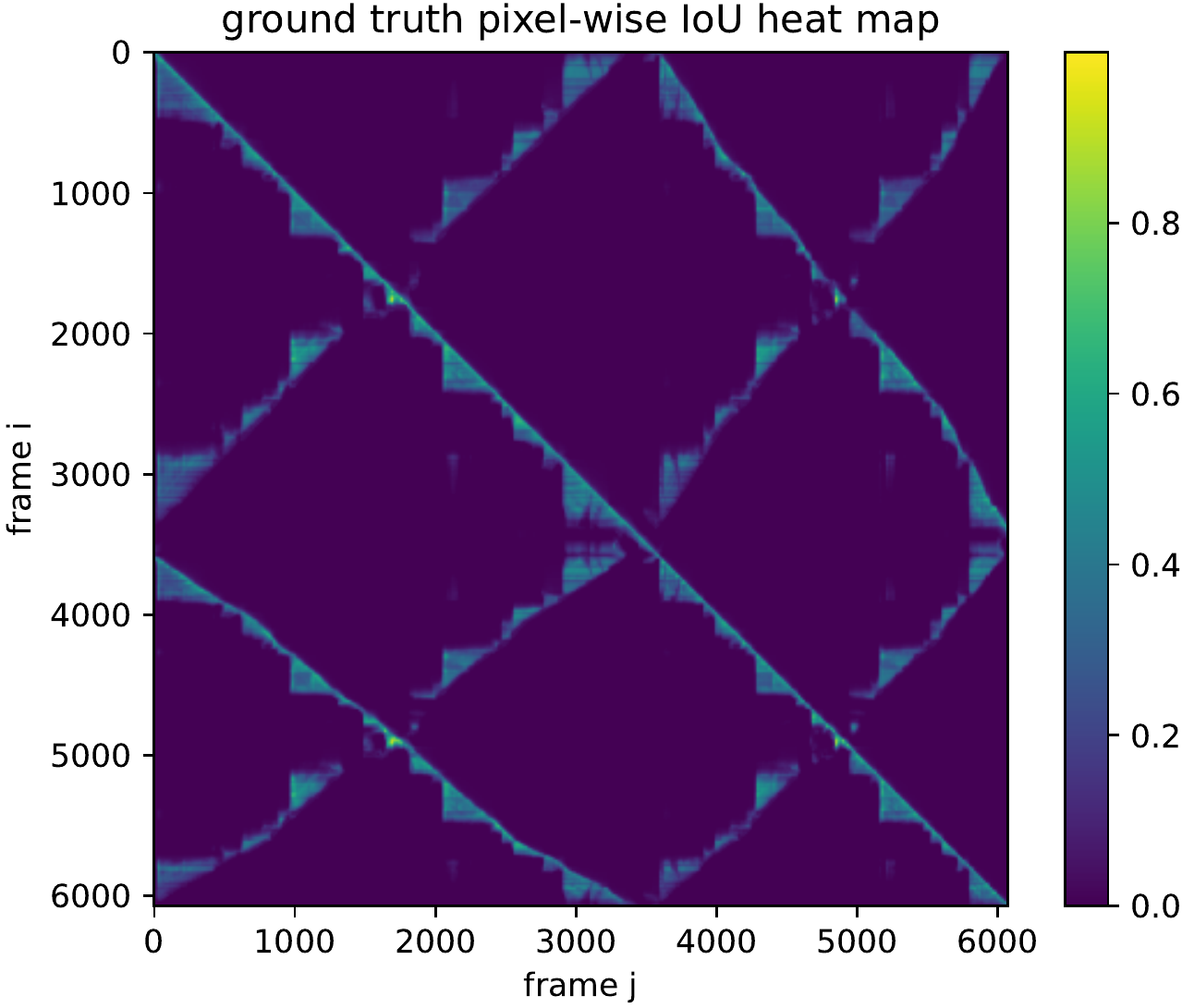}
  \caption{}
  \label{fig:reproj_iou_gt}
\end{subfigure}\hfil % <-- added
\begin{subfigure}{0.15\textwidth}
  \includegraphics[width=\linewidth]{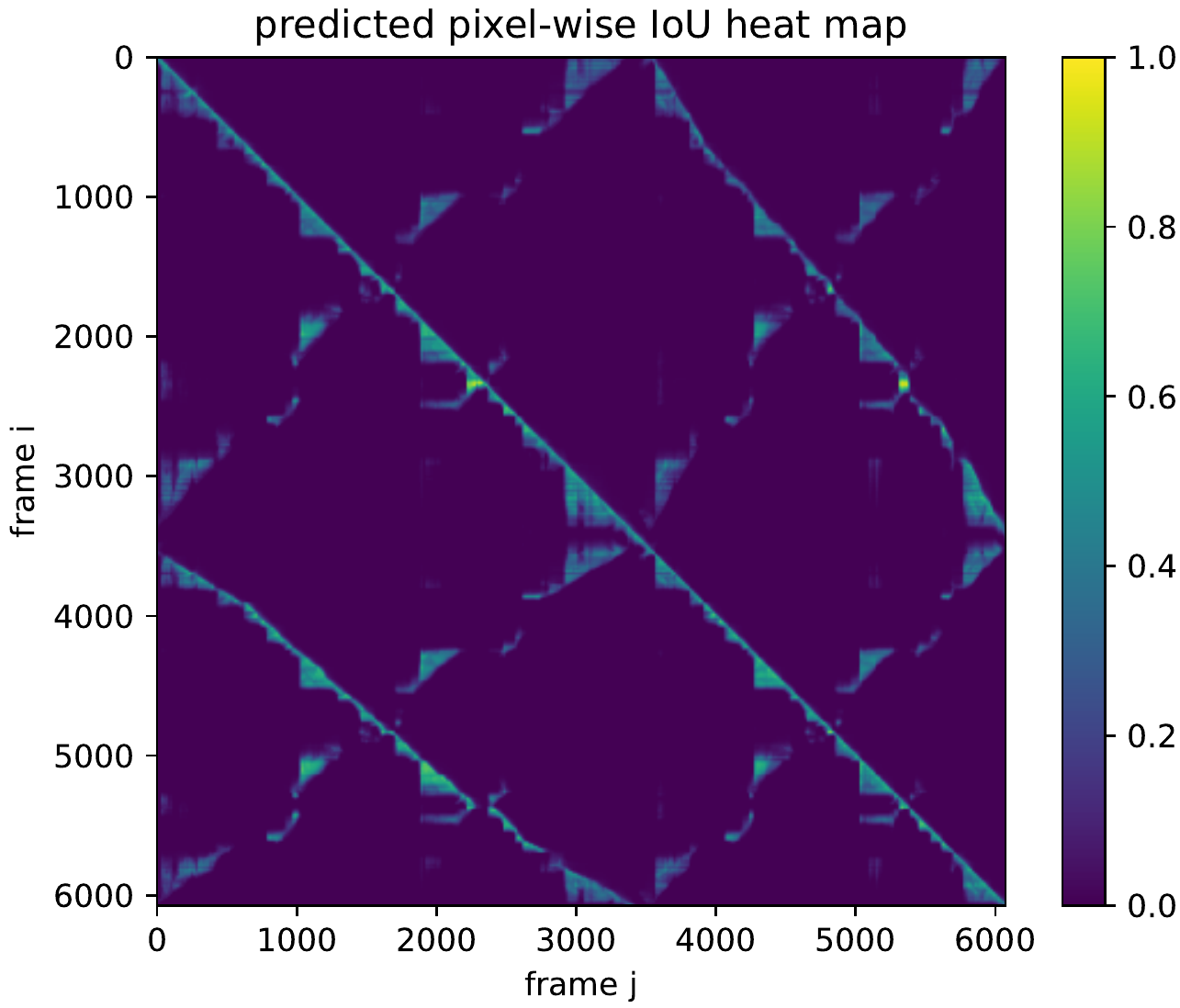}
  \caption{}
  \label{fig:reproj_iou_pred}
\end{subfigure}\hfil % <-- added
\begin{subfigure}{0.15\textwidth}
  \includegraphics[width=\linewidth]{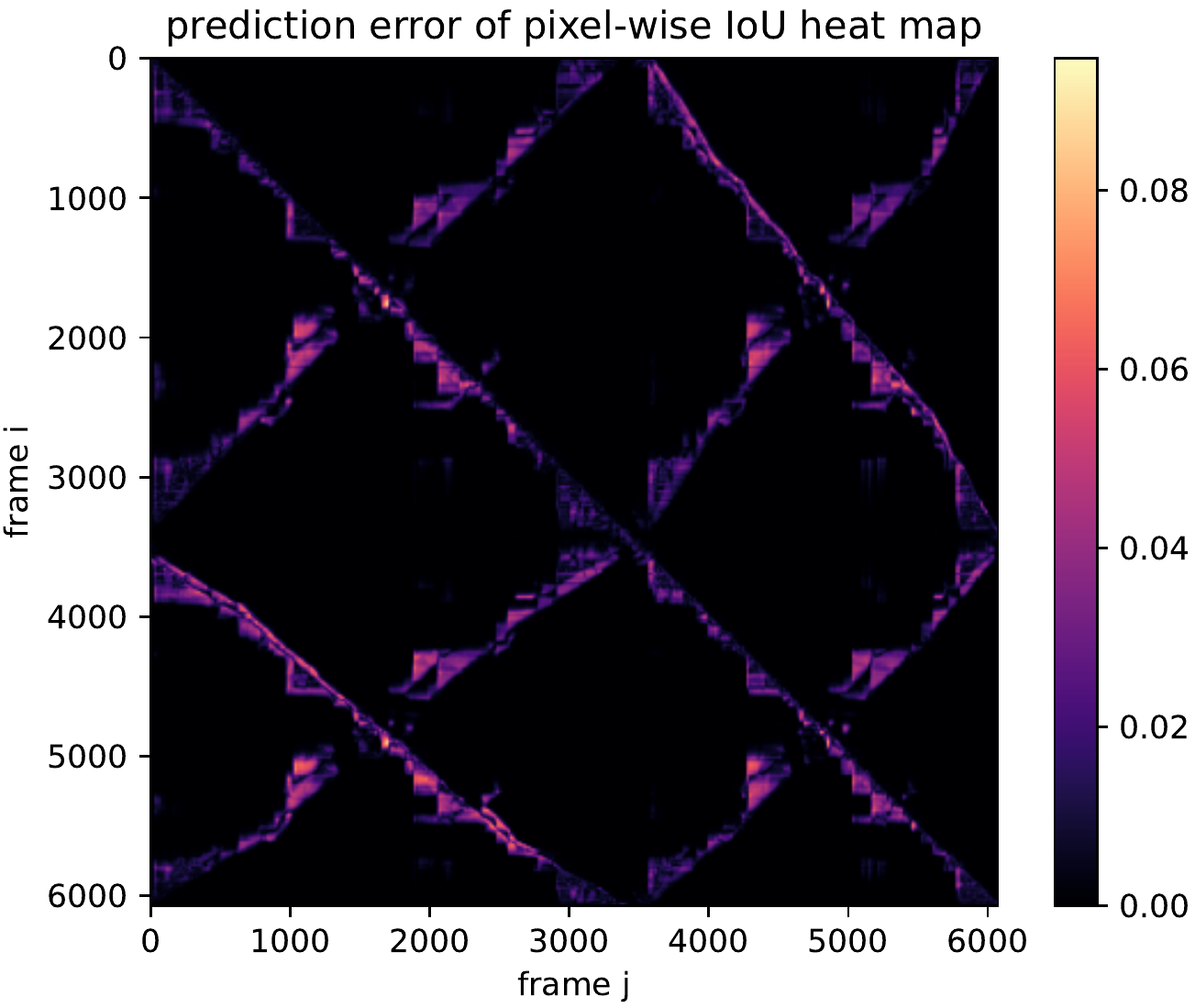}
  \caption{}
  \label{fig:reproj_error}
\end{subfigure}
\caption{Evaluation of similarity prediction. (a) ground truth similarity heatmap from reprojection IoU (b) predicted similarity heatmap from S3E (c) prediction error heatmap}
\label{fig:comparison_reproj_iou}
\end{figure}

In this section, we visualize the predicted similarity score from ViT powered S3E coding module in comparison with ground truth reprojection IoU similarity score in an untrained sequence. To better illustrate the performance of the scene embedding module, we provide the error map of the predicted similarity map. As shown in Fig. \ref{fig:comparison_reproj_iou}, the reprojection IoU heatmap describes how similar two pairs of images are in a tracked pose graph. The diagonal of the heatmap value is high since its self-projection. On the other hand, the off-diagonal highlighted parts indicate the reversed loop. Since we have applied the z-buffer method to remove occlusion, the similarity matrix is mostly symmetric. The error map proves that most activated similar areas align with the ground truth heatmap (where error exists), whereas the maximum absolute similarity error is less than 0.1. This means the predicted similarity, in spite of numeric errors, has implicitly learned the relative transformation and depth information for each pair of images.

\begin{figure}[htb]
    \centering % <-- added
\begin{subfigure}{0.23\textwidth}
  \includegraphics[width=\linewidth]{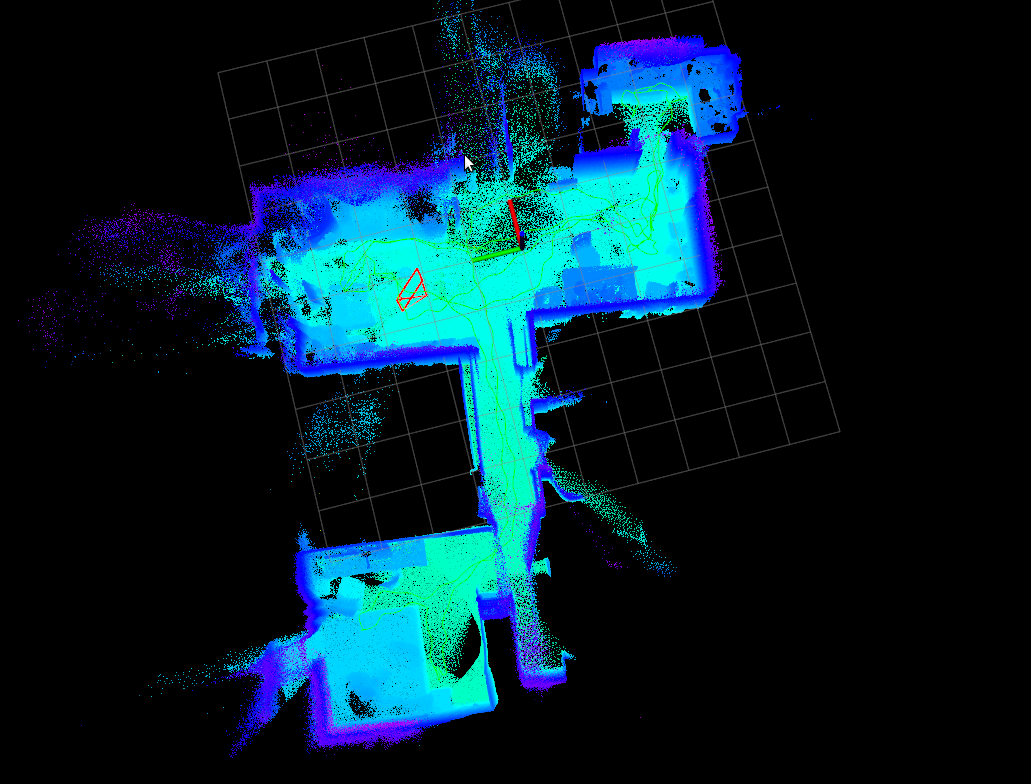}
  \caption{}
  \label{fig:map_traj_pcd}
\end{subfigure}\hfil % <-- added
\begin{subfigure}{0.23\textwidth}
  \includegraphics[width=\linewidth]{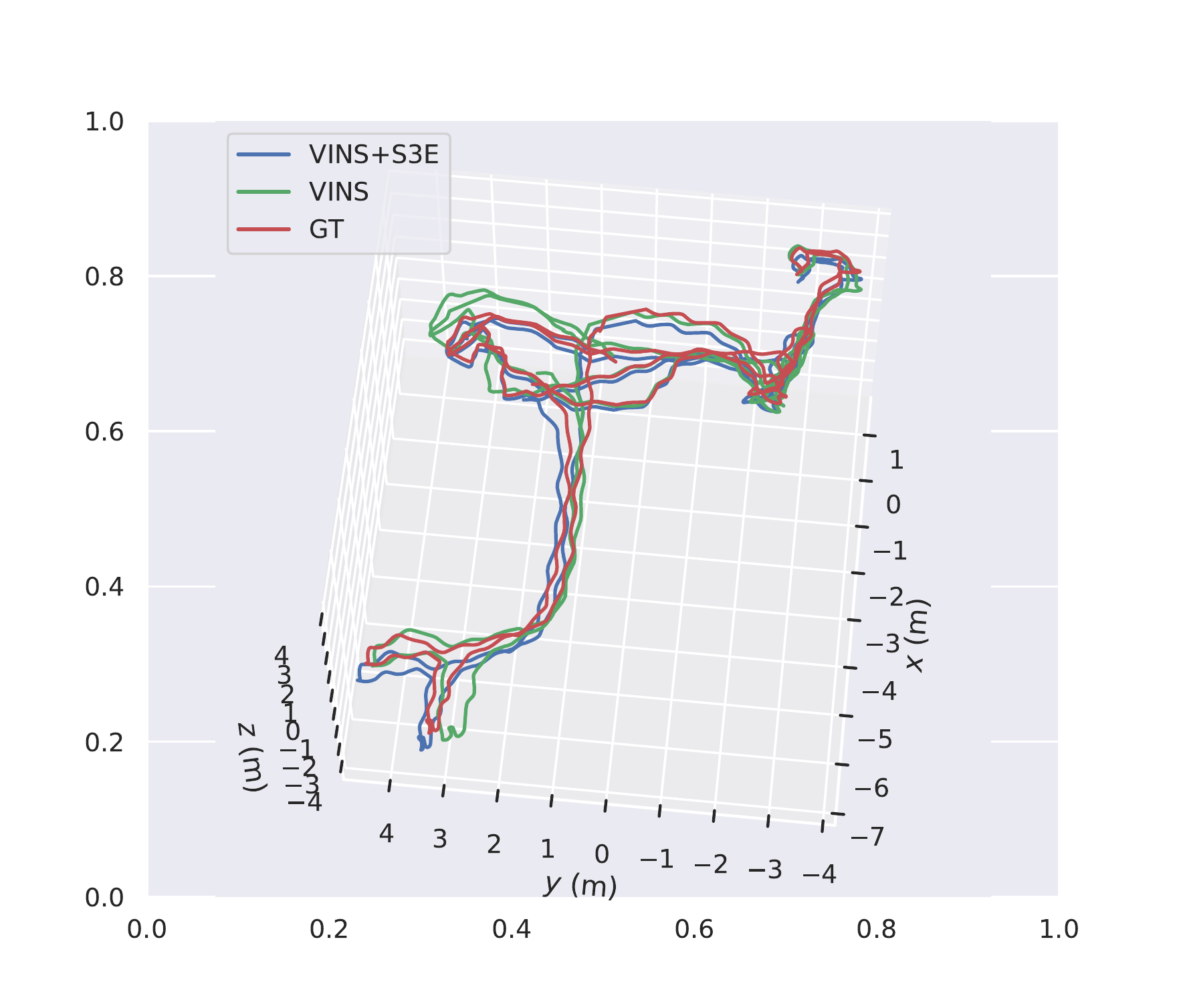}
  \caption{}
  \label{fig:traj_pose_vis_compare}
\end{subfigure}
\caption{Experiment environment 3D map and trajectory comparison.}
\label{fig:evaluation_scene}
\vspace{-10px}
\end{figure}

Moreover, we demonstrate the corresponding 3D scene with the estimated pose trajectory. Fig. \ref{fig:map_traj_pcd} shows the 3D point cloud map built from the VINS-Mono \cite{qin2018vins} with S3E-GNN. With the additional loop pairs detected by S3E-GNN, the estimated map is precise that the point cloud depicting the walls is as thin as 2 cm. Particularly, the pose comparison from Fig. \ref{fig:traj_pose_vis_compare} proves that the S3E-GNN improves the pose estimation and optimizes the global tracked pose graph. Trajectory estimation error by VINS-Mono with S3E-GNN is deducted by 41\%, compared to the one without S3E-GNN.

%===============================================================================
\section{FUTURE WORK}

Our S3E-GNN framework is currently trained on a dataset of indoor scenes and has demonstrated better performance than traditional BoW methods. In the future, we can extend the dataset to outdoor scenes, which will pave the way for applying our method to autonomous driving, drones, etc. Also, thanks to the flexible node addition and deletion strategies of our graph structure, we can use this method to deploy multiple cooperative robots. This application will play a crucial role in quickly collecting information over large areas, building environment models, and collecting situational awareness. 

\section{CONCLUSION}
\label{sec:conclusion}

In this paper, we introduce S3E-GNN, an end-to-end learning framework to improve the efficiency and robustness of camera relocalization. We propose a sparse spatial scene embedding module to encode the spatial and semantic features of a keyframe into a latent embedding code. By aggregating the embedding codes of all the keyframes into a reference pose graph, we can relocalize camera scenes using a GNN query algorithm. Our experiments demonstrate a notable improvement over the traditional BoW methods by implicitly incorporate the multi-view geometrical information with the additional scene matching constraints. We also show that our S3E-GNN framework can be employed as a plug-and-play module to any state-of-the-art SLAM systems.

% \newpage

% \addtolength{\textheight}{-12cm}   % This command serves to balance the column lengths
%                                   % on the last page of the document manually. It shortens
%                                   % the textheight of the last page by a suitable amount.
%                                   % This command does not take effect until the next page
%                                   % so it should come on the page before the last. Make
%                                   % sure that you do not shorten the textheight too much.

%%%%%%%%%%%%%%%%%%%%%%%%%%%%%%%%%%%%%%%%%%%%%%%%%%%%%%%%%%%%%%%%%%%%%%%%%%%%%%%%

%%%%%%%%%%%%%%%%%%%%%%%%%%%%%%%%%%%%%%%%%%%%%%%%%%%%%%%%%%%%%%%%%%%%%%%%%%%%%%%%

%%%%%%%%%%%%%%%%%%%%%%%%%%%%%%%%%%%%%%%%%%%%%%%%%%%%%%%%%%%%%%%%%%%%%%%%%%%%%%%%
% \section*{APPENDIX}

% Appendixes should appear before the acknowledgment.

% \section*{ACKNOWLEDGMENT}

% The preferred spelling of the word ÒacknowledgmentÓ in America is without an ÒeÓ after the ÒgÓ. Avoid the stilted expression, ÒOne of us (R. B. G.) thanks . . .Ó  Instead, try ÒR. B. G. thanksÓ. Put sponsor acknowledgments in the unnumbered footnote on the first page.

%%%%%%%%%%%%%%%%%%%%%%%%%%%%%%%%%%%%%%%%%%%%%%%%%%%%%%%%%%%%%%%%%%%%%%%%%%%%%%%%

% \medskip
% \bibliographystyle{IEEEtranN}
% \bibliographystyle{unsrtnat}
\bibliographystyle{IEEEtran}
\bibliography{citations}
\end{document}

\maketitle
\thispagestyle{empty}
\pagestyle{empty}

\section{Appendix A: Experiment Details}

\subsection{Proposed Model Details}

Additional details of our proposed model, which includes the attention pipeline and the reinforcement learning pipeline, and their corresponding training schemes are outlined below. 

\textbf{Attention Pipeline.} The attention pipeline is used to construct an informative attention map from a single color image in four main steps: semantic segmentation via U-Net segmentation network \cite{ronneberger2015u}, boundary extraction, boundary diffusion \cite{simoncini2016computational}, and inverse depth fusion (from ground-truth depth maps). Of these components, the U-Net segmentation network is the only learning based module and is pre-trained in simulator with the following scheme: 0.001 learning rate, 0.001 weight decay, 2e-5 decay rate, step scheduler, and cross-entropy loss function. Note that no gradient updates are made to the segmentation network during the reinforcement learning process, rendering the attention pipeline as fixed state representation module. However, we enable the RL agent to augment the state representations (i.e. attention maps) according to the task through attention augmentation as described in the paper. 

\begin{figure*}[htb!]
    \centering
    \includegraphics[width=0.9\textwidth]{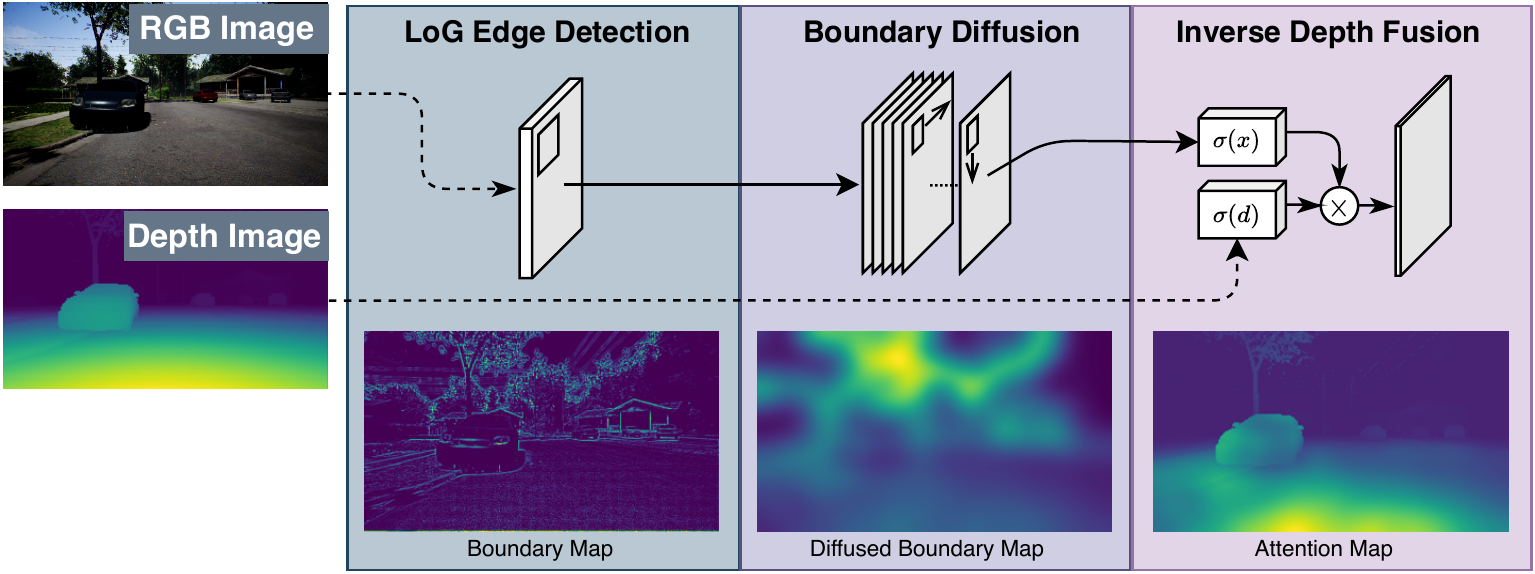}
    \caption{Unsupervised Attention Pipeline. Constructs an attention map from RGB image and depth map inputs using a Laplacian of Gaussian edge detector as opposed to semantic segmentation. Note that the boundaries are slightly noisier due to artifacts on the road which generates attention in those regions.}
    \vspace{-10px}
    \label{fig:att_depth_pipline_edge}
\end{figure*}

The unsupervised variant of the attention pipeline is shown in Fig. \ref{fig:att_depth_pipline_edge}, relying on a classical edge detection algorithm for locating object boundaries rather than segmentation. Note that this pipeline contains no learning-based components, and uses a fixed Laplacian of Gaussian (LoG) kernel (3x3) to produce the boundary map. Although the LoG edge detection algorithm is more susceptible to environmental noise (e.g. leaves, snow), our experimental results indicate that the unsupervised pipeline still strongly improves upon the end-to-end RL baselines and is comparable to our full model.

\begin{table}[htb]
\centering
\begin{tabular}{ccc}
\hline
\multicolumn{1}{|c|}{Layer (type)}       & \multicolumn{1}{c|}{Output Shape}            & \multicolumn{1}{c|}{Param \#}  \\ \hline
\multicolumn{1}{|c|}{Conv2d-1}           & \multicolumn{1}{c|}{{[}-1, 64, 256, 256{]}}  & \multicolumn{1}{c|}{1,792}     \\ \hline
\multicolumn{1}{|c|}{Conv2d-4}           & \multicolumn{1}{c|}{{[}-1, 64, 256, 256{]}}  & \multicolumn{1}{c|}{36,928}    \\ \hline
\multicolumn{1}{|c|}{Conv2d-10}          & \multicolumn{1}{c|}{{[}-1, 128, 128, 128{]}} & \multicolumn{1}{c|}{73,856}    \\ \hline
\multicolumn{1}{|c|}{Conv2d-13}          & \multicolumn{1}{c|}{{[}-1, 128, 128, 128{]}} & \multicolumn{1}{c|}{147,584}   \\ \hline
\multicolumn{1}{|c|}{Conv2d-19}          & \multicolumn{1}{c|}{{[}-1, 256, 64, 64{]}}   & \multicolumn{1}{c|}{295,168}   \\ \hline
\multicolumn{1}{|c|}{Conv2d-22}          & \multicolumn{1}{c|}{{[}-1, 256, 64, 64{]}}   & \multicolumn{1}{c|}{590,080}   \\ \hline
\multicolumn{1}{|c|}{Conv2d-28}          & \multicolumn{1}{c|}{{[}-1, 512, 32, 32{]}}   & \multicolumn{1}{c|}{1,180,160} \\ \hline
\multicolumn{1}{|c|}{Conv2d-31}          & \multicolumn{1}{c|}{{[}-1, 512, 32, 32{]}}   & \multicolumn{1}{c|}{2,359,808} \\ \hline
\multicolumn{1}{|c|}{Conv2d-37}          & \multicolumn{1}{c|}{{[}-1, 512, 16, 16{]}}   & \multicolumn{1}{c|}{2,359,808} \\ \hline
\multicolumn{1}{|c|}{Conv2d-40}          & \multicolumn{1}{c|}{{[}-1, 512, 16, 16{]}}   & \multicolumn{1}{c|}{2,359,808} \\ \hline
\multicolumn{1}{|c|}{ConvTranspose2d-45} & \multicolumn{1}{c|}{{[}-1, 512, 32, 32{]}}   & \multicolumn{1}{c|}{1,049,088} \\ \hline
\multicolumn{1}{|c|}{Conv2d-46}          & \multicolumn{1}{c|}{{[}-1, 256, 32, 32{]}}   & \multicolumn{1}{c|}{2,359,552} \\ \hline
\multicolumn{1}{|c|}{Conv2d-49}          & \multicolumn{1}{c|}{{[}-1, 256, 32, 32{]}}   & \multicolumn{1}{c|}{590,080}   \\ \hline
\multicolumn{1}{|c|}{ConvTranspose2d-54} & \multicolumn{1}{c|}{{[}-1, 256, 64, 64{]}}   & \multicolumn{1}{c|}{262,400}   \\ \hline
\multicolumn{1}{|c|}{Conv2d-55}          & \multicolumn{1}{c|}{{[}-1, 128, 64, 64{]}}   & \multicolumn{1}{c|}{589,952}   \\ \hline
\multicolumn{1}{|c|}{Conv2d-58}          & \multicolumn{1}{c|}{{[}-1, 128, 64, 64{]}}   & \multicolumn{1}{c|}{147,584}   \\ \hline
\multicolumn{1}{|c|}{ConvTranspose2d-63} & \multicolumn{1}{c|}{{[}-1, 128, 128, 128{]}} & \multicolumn{1}{c|}{65,664}    \\ \hline
\multicolumn{1}{|c|}{Conv2d-64}          & \multicolumn{1}{c|}{{[}-1, 64, 128, 128{]}}  & \multicolumn{1}{c|}{147,520}   \\ \hline
\multicolumn{1}{|c|}{Conv2d-67}          & \multicolumn{1}{c|}{{[}-1, 64, 128, 128{]}}  & \multicolumn{1}{c|}{36,928}    \\ \hline
\multicolumn{1}{|c|}{ConvTranspose2d-72} & \multicolumn{1}{c|}{{[}-1, 64, 256, 256{]}}  & \multicolumn{1}{c|}{16,448}    \\ \hline
\multicolumn{1}{|c|}{Conv2d-73}          & \multicolumn{1}{c|}{{[}-1, 64, 256, 256{]}}  & \multicolumn{1}{c|}{73,792}    \\ \hline
\multicolumn{1}{|c|}{Conv2d-76}          & \multicolumn{1}{c|}{{[}-1, 64, 256, 256{]}}  & \multicolumn{1}{c|}{36,928}    \\ \hline
\multicolumn{1}{|c|}{Conv2d-81}          & \multicolumn{1}{c|}{{[}-1, 1, 256, 256{]}}   & \multicolumn{1}{c|}{65}        \\ \hline
\multicolumn{3}{l}{Total params: 14,788,929}                                                                             \\
\multicolumn{3}{l}{Trainable params: 14,788,929}                                                                 
\end{tabular}
\caption{Unsupervised encoder-decoder model architecture.}
\label{table: unsupervised_enc_dec_model_architecture}
\end{table}
\textbf{Reinforcement Learning Pipeline.} Our proposed reinforcement learning model follows the DQN \cite{mnih2013playing} setup, and features several key extensions - the most prominent of them being the integration with a fixed state attention module, but also a unique configuration of neural networks that promotes fast, stable, and robust self-driving behaviour learning. Network architecture specifics and hyperparameters are presented in Table. \ref{table: lstm_dqn_architecture}.

% \begin{table}[htb]
% \centering
% \begin{tabular}{|l|l|l|l|l|}
% \hline
% Layers & Layers & input shape 84x84      & input shape 32x32      & input shape 16x16      \\ \hline
% \multirow{4}{*}{Conv}   & \multirow{3}{*}{Conv}   & conv1(1, 32, 8, 1, 1)         & conv1(1, 32, 5, 1, 1)        & conv1(1, 32, 3, 1, 1)       \\ \cline{3-5} 
%       &        & conv2(32, 64, 5, 1, 1) & conv2(32, 64, 3, 1, 1) & conv2(32, 64, 1, 1, 1) \\ \cline{3-5} 
%       &        & conv3(64, 64, 3, 1, 1) & conv3(64, 64, 1, 1, 1) & conv3(64, 64, 1, 1, 1) \\ \cline{2-5} 
%       & output & {[}64, 4, 4{]}         & {[}64, 4, 4{]}         & {[}64, 4, 4{]}         \\ \hline
% \multirow{3}{*}{LSTM}   & \multirow{2}{*}{LSTM}   & \multicolumn{3}{l|}{{[}32, 192, 16{]} (window\_size, concated\_features, feature\_vector)}  \\ \cline{3-5} 
%       &        & \multicolumn{3}{l|}{LSTM(16, 128)}                                       \\ \cline{2-5} 
%                         & output                  & \multicolumn{3}{l|}{{[}32, 192, 128{]} (window\_size, concated\_features, feature\_vector)} \\ \hline
% \multirow{3}{*}{Linear} & \multirow{2}{*}{Linear} & \multicolumn{3}{l|}{Linear(24576, 512)}                                                    \\ \cline{3-5} 
%       &        & \multicolumn{3}{l|}{Linear(512, 2)}                                      \\ \cline{2-5} 
%       & output & \multicolumn{3}{l|}{{[}32, 2{]}}                                         \\ \hline
% \end{tabular}
% \caption{LSTM DQN Architecture}
% \end{table}

\begin{table}[htb]
\centering
\resizebox{\columnwidth}{!}{
\begin{tabular}{ccccc}
\hline
\multicolumn{1}{|l|}{Layers} &
  \multicolumn{1}{l|}{Layers} &
  \multicolumn{1}{l|}{input shape 84x84} &
  \multicolumn{1}{l|}{input shape 32x32} &
  \multicolumn{1}{l|}{input shape 16x16} \\ \hline
\multicolumn{1}{|l|}{\multirow{4}{*}{Conv}} &
  \multicolumn{1}{l|}{\multirow{3}{*}{Conv}} &
  \multicolumn{1}{l|}{conv1(1, 32, 8, 1, 1)} &
  \multicolumn{1}{l|}{conv1(1, 32, 5, 1, 1)} &
  \multicolumn{1}{l|}{conv1(1, 32, 3, 1, 1)} \\ \cline{3-5} 
\multicolumn{1}{|l|}{} &
  \multicolumn{1}{l|}{} &
  \multicolumn{1}{l|}{conv2(32, 64, 5, 1, 1)} &
  \multicolumn{1}{l|}{conv2(32, 64, 3, 1, 1)} &
  \multicolumn{1}{l|}{conv2(32, 64, 1, 1, 1)} \\ \cline{3-5} 
\multicolumn{1}{|l|}{} &
  \multicolumn{1}{l|}{} &
  \multicolumn{1}{l|}{conv3(64, 64, 3, 1, 1)} &
  \multicolumn{1}{l|}{conv3(64, 64, 1, 1, 1)} &
  \multicolumn{1}{l|}{conv3(64, 64, 1, 1, 1)} \\ \cline{2-5} 
\multicolumn{1}{|l|}{} &
  \multicolumn{1}{l|}{output} &
  \multicolumn{1}{l|}{{[}64, 4, 4{]}} &
  \multicolumn{1}{l|}{{[}64, 4, 4{]}} &
  \multicolumn{1}{l|}{{[}64, 4, 4{]}} \\ \hline
\multicolumn{1}{|l|}{\multirow{3}{*}{LSTM}} &
  \multicolumn{1}{l|}{\multirow{2}{*}{LSTM}} &
  \multicolumn{3}{l|}{{[}32, 192, 16{]} (window\_size, concated\_features, feature\_vector)} \\ \cline{3-5} 
\multicolumn{1}{|l|}{} &
  \multicolumn{1}{l|}{} &
  \multicolumn{3}{l|}{LSTM(16, 128)} \\ \cline{2-5} 
\multicolumn{1}{|l|}{} &
  \multicolumn{1}{l|}{output} &
  \multicolumn{3}{l|}{{[}32, 192, 128{]} (window\_size, concated\_features, feature\_vector)} \\ \hline
\multicolumn{1}{|l|}{\multirow{3}{*}{Linear}} &
  \multicolumn{1}{l|}{\multirow{2}{*}{Linear}} &
  \multicolumn{3}{l|}{Linear(24576, 512)} \\ \cline{3-5} 
\multicolumn{1}{|l|}{} &
  \multicolumn{1}{l|}{} &
  \multicolumn{3}{l|}{Linear(512, 2)} \\ \cline{2-5} 
\multicolumn{1}{|l|}{} &
  \multicolumn{1}{l|}{output} &
  \multicolumn{3}{l|}{{[}32, 2{]}} \\ \hline
\multicolumn{5}{l}{Learning rate: 0.0001, play\_interval: 900, target\_update\_interval: 1000} \\
\multicolumn{5}{l}{Replay memory: 50000, epsilon\_start: 1, epsilon\_end: 0.01, epsilon\_decay: 100000}
\end{tabular}
}
\caption{LSTM DQN architecture and hyper-parameters.}
\label{table: lstm_dqn_architecture}
\end{table}

\subsection{Baseline Model Details}

\begin{table}[htb]
\resizebox{\columnwidth}{!}{
\begin{tabular}{cccccc}
\hline
\multicolumn{3}{|c|}{DQN} & \multicolumn{3}{c|}{DDPG} \\ \hline
\multicolumn{1}{|c|}{Layer (type)} & \multicolumn{1}{c|}{Output Shape} & \multicolumn{1}{c|}{Param \#} & \multicolumn{1}{c|}{Layer (type)} & \multicolumn{1}{c|}{Output Shape} & \multicolumn{1}{c|}{Param \#} \\ \hline
\multicolumn{1}{|c|}{Conv2d(4, 32,8,4,0)} & \multicolumn{1}{c|}{{[}-1, 32, 20, 20{]}} & \multicolumn{1}{c|}{8,224} & \multicolumn{1}{c|}{Conv2d(4, 32,8,4,0)} & \multicolumn{1}{c|}{{[}-1, 32, 40, 40{]}} & \multicolumn{1}{c|}{3,232} \\ \hline
\multicolumn{1}{|c|}{Conv2d(32,64, 4,2,0)} & \multicolumn{1}{c|}{{[}-1, 64, 9, 9{]}} & \multicolumn{1}{c|}{32,832} & \multicolumn{1}{c|}{Conv2d(32,64, 4,2,0)} & \multicolumn{1}{c|}{{[}-1, 64, 38, 38{]}} & \multicolumn{1}{c|}{18,496} \\ \hline
\multicolumn{1}{|c|}{Conv2d(64, 64, 3, 1, 0)} & \multicolumn{1}{c|}{{[}-1, 64, 7, 7{]}} & \multicolumn{1}{c|}{36,928} & \multicolumn{1}{c|}{Conv2d(64, 64, 3, 1, 0)} & \multicolumn{1}{c|}{{[}-1, 64, 38, 38{]}} & \multicolumn{1}{c|}{4,160} \\ \hline
\multicolumn{1}{|c|}{Linear(3236, 512)} & \multicolumn{1}{c|}{{[}-1, 512{]}} & \multicolumn{1}{c|}{1,606,144} & \multicolumn{1}{c|}{Linear(92416+1, 512)} & \multicolumn{1}{c|}{{[}-1, 512{]}} & \multicolumn{1}{c|}{47,318,016} \\ \hline
\multicolumn{1}{|c|}{Linear(512, 2)} & \multicolumn{1}{c|}{{[}-1, 2{]}} & \multicolumn{1}{c|}{1,026} & \multicolumn{1}{c|}{Linear(512, 2)} & \multicolumn{1}{c|}{{[}-1, 2{]}} & \multicolumn{1}{c|}{1,026} \\ \hline
\multicolumn{3}{|c|}{Total params: 1,685,154} & \multicolumn{3}{c|}{Total params: 47,344,930} \\ \hline
\multicolumn{3}{|c|}{Trainable params: 1,685,154} & \multicolumn{3}{c|}{Trainable params: 47344930} \\ \hline
\multicolumn{3}{|c|}{\multirow{3}{*}{Please Refer to Table \ref{table: lstm_dqn_architecture}}} & \multicolumn{3}{c|}{Buffer size: 100000, batch size: 32, $\gamma$: 0.99} \\ \cline{4-6} 
\multicolumn{3}{|c|}{} & \multicolumn{3}{c|}{$\tau$: 0.001, LRA: 0.0001, LRC: 0.001} \\ \cline{4-6} 
\multicolumn{3}{|c|}{} & \multicolumn{3}{c|}{Explore: 1,000,000} \\ \hline
\multicolumn{1}{|l|}{Reward} & \multicolumn{2}{l|}{0.915} & \multicolumn{1}{l|}{Reward} & \multicolumn{2}{l|}{1.357} \\ \hline
\multicolumn{6}{l}{$\tau$: target network hyper-parameter} \\
\multicolumn{6}{l}{LRA: learning rate for actor network} \\
\multicolumn{6}{l}{LRC: learning rate for critic network}
\end{tabular}
}
\caption{DQN and DDPG model architecture and hyper-parameters.}
\label{table: dqn_ddpg_model_architecture}
\end{table}

Since our model is based off the DQN framework, we benchmark it's performance against the traditional DQN \cite{mnih2013playing} framework and DDPG \cite{lillicrap2015continuous} that features actor and critic networks. Both of these models learn directly from color images (end-to-end) and are updated by the same loss function and sparse rewards as our proposed method for consistency; see Table. \ref{table: dqn_ddpg_model_architecture} for architecture and training details. We extend these two baselines by integrating them with the attention pipeline and employing attention augmentation with the same unsupervised encoder/decoder network; thus, comparison highlights the effect of our multi-scale latent encoding and LSTM Q-network in our method. We observe a significant improvement to the lower-bound convergence of our baselines after incorporating attention augmentation. Also, a substantial increase in data efficiency and reduced variance is observed when training our method. The results of all experiments are averaged over 7 runs. 

\medskip
\bibliographystyle{IEEEtran}
\bibliography{citations}